\documentclass{sig-alternate}

\pdfoutput=1

\newfont{\mycrnotice}{ptmr8t at 7pt}
\newfont{\myconfname}{ptmri8t at 7pt}
\let\crnotice\mycrnotice%
\let\confname\myconfname%

\permission{Permission to make digital or hard copies of all or part of this work for personal or classroom use is granted without fee provided that copies are not made or distributed for profit or commercial advantage and that copies bear this notice and the full citation on the first page. Copyrights for components of this work owned by others than the author(s) must be honored. Abstracting with credit is permitted. To copy otherwise, or republish, to post on servers or to redistribute to lists, requires prior specific permission and/or a fee. Request permissions from permissions@acm.org.}
\conferenceinfo{UnstructureNLP'13,}{October 28, 2013, San Francisco, CA, USA. \\
{\mycrnotice{Copyright is held by the owner/author(s). Publication rights licensed to ACM.}}}
\copyrightetc{ACM \the\acmcopyr\ ...\$15.00}
\crdata{978-1-4503-2415-1/13/10 \\
http://dx.doi.org/10.1145/2513549.2513558}

\usepackage[hide]{chato-notes}

\usepackage[hyphens]{url} 
\usepackage{paralist}
\usepackage{subfigure}
\usepackage{booktabs}
\usepackage[bf,tableposition=top]{caption}
\usepackage{multirow}
\usepackage{epstopdf}
\usepackage{array}
\usepackage{graphicx}
\graphicspath{{img/}}
\usepackage[numbers, square]{natbib}
\usepackage[bookmarks, pdftex, colorlinks=false]{hyperref}
\usepackage{xspace}

\newcommand{\spara}[1]{\smallskip\noindent{\bf #1}}
\newcommand{\mpara}[1]{\medskip\noindent{\bf #1}}

\newenvironment {squishlist}
{\begin{list}{$\bullet$}
  { \setlength{\itemsep}{0pt}
     \setlength{\parsep}{3pt}
     \setlength{\topsep}{3pt}
     \setlength{\partopsep}{0pt}
     \setlength{\leftmargin}{1.5em}
     \setlength{\labelwidth}{1em}
     \setlength{\labelsep}{0.5em} } }
{\end{list}}

\newcommand{\aGeneral}{\texttt{gen}}
\newcommand{\aSports}{\texttt{spt}}
\newcommand{\aEntertainment}{\texttt{ent}}
\newcommand{\aBusiness}{\texttt{biz}}
\newcommand{\fGeneral}{\texttt{general}\xspace}
\newcommand{\fSports}{\texttt{sports}\xspace}
\newcommand{\fEntertainment}{\texttt{entertainment}\xspace}
\newcommand{\fBusiness}{\texttt{business}\xspace}

\newcommand{\provider}[2]{{\tt #1[#2]}}
\newcommand{\prov}[1]{{\tt #1}}

\newcounter{cntQuestion} 
\newenvironment{question}[1][]{\refstepcounter{cntQuestion}\smallskip\noindent {\textbf{Q\thecntQuestion: #1.} }{}}{}

\title{Says who? Automatic Text-Based\\Content Analysis of Television News}

\begin{document}

\numberofauthors{4}
\author{
%% 1st. author
\alignauthor
Carlos Castillo\\
       \affaddr{Qatar Computing Research Institute}\\
       \affaddr{Doha, Qatar}\\
       \email{chato@acm.org}
%% 2nd. author
\alignauthor
Gianmarco De~Francisci~Morales\\
       \affaddr{Yahoo! Research}\\
       \affaddr{Barcelona, Spain}\\
       \email{gdfm@yahoo-inc.com}
%% 3rd. author
\alignauthor
Marcelo Mendoza\\
       \affaddr{Yahoo! Research}\\
       \affaddr{Santiago, Chile}\\
       \email{mendozam@yahoo-inc.com}
%% 3rd. author
\and
\alignauthor
Nasir Khan\\
       \affaddr{Al Jazeera}\\
       \affaddr{Doha, Qatar}\\
       \email{nasir.khan@aljazeera.net}
}
\maketitle

%!TEX root = paper.tex
% Speak about text, evaluation
\begin{abstract}
We perform an automatic analysis of television news programs, based on the closed captions that accompany them. Specifically, we collect all the news broadcasted in over 140 television channels in the US during a period of six months. We start by segmenting, processing, and annotating the closed captions automatically. Next, we focus on the analysis of their linguistic style and on mentions of people using NLP methods. We present a series of key insights about news providers, people in the news, and we discuss the biases that can be uncovered by automatic means. These insights are contrasted by looking at the data from multiple points of view, including qualitative assessment.
\end{abstract}

\section{Introduction}\label{sec:intro}

Television is a dominant source of news today, wielding an enormous influence over many aspects of our life. 
The ascent of the Web has caused a significant drop in newspaper and radio audiences, but television remains the number one source for news in the US~\cite{pew2012tv}.

We analyze the closed captions of newscasts, which are provided by the news networks themselves.
By using these streams of text, we study how to characterize each news network, each person-type named entity mentioned in the news ({\em newsmaker}), and the relationship between news networks and newsmakers (e.g., the biases of networks in the coverage of news related to a person).
To the best of our knowledge, this work is the first to perform text analysis of content in television news with such broad goals and in such an ambitious scale. % broad goals = many different things, ambitious scale = 6 months of data; I would like to say large scale but it's only 2MB of text per day --ChaTo

We introduce a NLP-based pipeline of tools to process the input stream of data.
While the specific choice of each tool is task-dependent, the pipeline itself and its components represent a minimum number of necessary steps to extract useful information from the data.
Any other data mining task that uses closed captions, such as text segmentation, classification, or clustering, can build upon a pre-processing pipeline similar to the one described here.

Our NLP-based pipeline filters out non-news programs, segments the captions into sentences, detects named entities (specifically people), applies a part-of-speech tagger to find words and qualifiers used together with each entity, and labels automatically each sentence with an overall sentiment score.

These tools extract a set of measurable dimensions from the text, which we employ to tackle the following tasks:

\begin{squishlist}
 \item characterize news providers and news programs in terms of style, news coverage and timeliness (Section~\ref{sec:providers});
 \item characterize newsmakers with respect to their popularity, profession and similarity to other people (Section~\ref{sec:newsmakers});
 \item characterize biases and framing in the coverage of newsmakers from different providers (Section~\ref{sec:biases}).
\end{squishlist}

%The next section outlines previous works related to ours. Section~\ref{sec:framework} introduces our dataset.
%%Sections~\ref{sec:providers} and~\ref{sec:newsmakers} describe the characterization of the different news providers and newsmakers, respectively, and Section~\ref{sec:biases} shows the interactions between them.
%Sections~\ref{sec:providers}, \ref{sec:newsmakers} and \ref{sec:biases} present our main results.
%Section~\ref{sec:conclusions} details our conclusions.

%!TEX root = paper.tex
\section{Related Work} \label{sec:relwork}

%The interest of the information retrieval community in television content is not new.
One of the oldest references on mining television content is a DARPA-sponsored workshop in 1999 with a topic detection and tracking challenge~\cite{cieri1999tdt}.
%The interest on this topic has been fueled by the need of indexing and searching large digital libraries of multimedia content, e.g., see the work by Smeaton et al.~\citep{smeaton2005large, smeaton_2003_segmentation}.
Higher-level applications have been emerging in recent years.
For example, \citet{henzinger_2006_news} describe a system for finding web pages related to television content, and test different methods to synthesize a web search query from a television transcript.
\citet{oger_2010_transcription} classify videos based on a transcription obtained from speech recognition.
\citet{xu_2011_credibility} describe a system to rate the credibility of information items on television by looking at how often the same image is described in a similar way by more than one news source.

Information from closed captions has also been combined with other data sources.
\citet{lin_2002_video_classification} present a system for video classification based on closed captions as well as content-based attributes of the video.
\citet{misra_2010_segmentation} also combine closed captions and multimedia content to improve video segmentation.
\citet{shamma_2009_debates} align the closed captions during a live event (a presidential debate) with social media reactions.
\citet{gibbon_2012_combining} combine closed captions with real-time television audience measures to detect ads -- which typically are accompanied by sudden drops in viewership (``{\em zapping}'').

\spara{Quantitative analysis of media.}
%Several research groups have used quantitative evidence to, among other tasks, measure the amount of bias of different news sources. For instance, \citet{alessio_2000_bias} study gatekeeping, coverage, and statement bias by annotating content from magazines and television in the US.
%They find no substantial bias on magazines in the US, and a small coverage bias on US television. % (52\% news about Democrats, 47\% news about Republicans). 
%\citet{schoenbach_2001_politicians} study Dutch and German television and observe that top leaders such as chancellors or prime ministers get a substantially larger %(3 to 8 times)
%number of mentions than the second most mentioned politicians.
%\citet{morris_2005_news} focus on US party conventions, %, manually coding samples of transcripts of news programs that cover each convention.
%They notice, among several other findings, that
%and notice that what people hear on the news is mostly commentary, with limited amount of quotes from the convention participants.
%Moreover, they find large differences in the sentiment polarity with which candidates are treated.% by different networks.

Most content analysis of news reports is based on a time-consuming process of manual coding;
automatic methods are less frequently used.
\citet{groseclose_2005_bias} use an indirect measure of television bias by manually counting references to think tanks in each network, and then by scoring each think tank on a left-right political spectrum by automatically counting their appearances in congressional records of speeches by politicians of different political leaning.
\citet{flaounas2013digital_journalism} study news articles available on the web, and analyze the prevalence of different topics and the distribution of readability and subjectivity scores. % of the text.

\spara{Topics and perspectives.}
The PhD thesis of~\citet{lin_2008_perspectives} and related papers (e.g. \citep{lin_2006_side}), introduce a joint probabilistic model of the topics and the perspectives from which documents and sentences are written. The parameters of the model are learned from a training corpus for which the ideological perspectives are known. In contrast, the methods we study on this paper are based on NLP algorithms.

Lin et al. process news videos using content-based methods. First, news boundaries are detected automatically using visual clues, such as scene and background changes. Second, each news segment is annotated with concepts such as {\em sky}, {\em person}, {\em outdoor}, etc. which are inferred automatically from shots in the video. This approach is effective for settings in which closed captions are not available.

% !TEX root = paper.tex
%\section{Data and tools} \label{sec:framework}

%This section introduces the data sources and data preparation steps we perform.

%\subsection{Data acquisition} \label{subsec:data}

\subsection{Text pre-processing} \label{subsec:tools}

We use closed captions provided by a software developed by us and recently presented in the software demonstration session of SIGIR~\cite{chato2013}.

%%software company that develops {\em second-screen experiences}, i.e., applications that display extra information about a TV show on a smartphone or tablet.
%The same data is also publicly available from the Internet Archive\footnote{\url{http://archive.org/details/tv}}, which recently opened a TV search service and whose archives date back to 2009.
We collected data that consist of all the closed captions from January 2012 to June 2012 for
about 140 channels. On average, each TV channel generates $\approx2$MB of text every day.

The closed captions are streams of plain text that we process through a series of steps. First, to segment the text stream into sentences we use a series of heuristics which include detecting a change of speaker, conventionally signaled by a text marker (``$\texttt{>}\texttt{>}$''), using the presence of full stops, and using time-based rules.
%For instance, a pause of several seconds indicates the presence of a new sentence.
We remark that there exist methods to join sentences into passages~\cite{smeaton_2003_segmentation,misra_2010_segmentation},
but for our analysis we use single sentences as basic units of content, and we only group them when they match to the same news item, as described in Section~\ref{subsec:news_matching}.

Second, we recognize and extract named entities by using a named entity tagger that works in two steps: entity resolution~\cite{zhou2010resolving} and ``aboutness'' ranking~\cite{paranjpe2009learning}.
We focus on the \textbf{person} type in the remainder of this paper, 
and whenever we find a given entity in the closed captions of a news provider, we count a \textbf{mention} of that person by the provider.

Third, we apply the Stanford NLP tagger~\cite{toutanova_2003_stanford_nlp} to perform part-of-speech tagging and dependency parsing.\footnote{
Further details on the tag set are available in the manual {\scriptsize{\url{http://nlp.stanford.edu/software/dependencies_manual.pdf}}}}
As a last step of the text pre-processing, we apply sentiment analysis to each sentence by using SentiStrength~\cite{thelwall2010sentistrength}.

\mpara{Example.}
A brief example can illustrate the main parts of our text pre-processing. The input data is similar to this:

{\small\begin{verbatim}
[1339302660.000]    WHAT MORE CAN YOU ASK FOR? 
[1339302662.169]    >> THIS IS WHAT NBA
[1339302663.203]    BASKETBALL IS ABOUT.
\end{verbatim}}

The TV channel maps to a network name (\prov{CNN}), and the time is used to look-up in the programming guide to determine the type ({\em CNN World Sports}, which is about \fSports news). Hence, this news provider is identified as \provider{CNN}{\aSports}. Finally, the text is tagged to generate the following output:

{\small\begin{verbatim}
What/WP more/JJR can/MD you/PRP ask/VB for/IN ?/. This/DT 
 is/VBZ what/WDT NBA/NNP [entity: National_Basketball_
 Association] basketball/NN is/VBZ about/IN ./.
\end{verbatim}}

\subsection{News matching}\label{subsec:news_matching}

We match the processed captions to recent news stories, which are obtained from a major online news aggregator. %that uses hundreds of different sources.
Captions are matched in the same genre, e.g., sentences in \fSports are matched to online news in the \fSports section of the news aggregator.
News in the website that are older than three days are ignored.
The matching task is the same as the one described by~\citet{henzinger_2006_news}, but the approach is based on supervised learning rather than web searches.%\footnote{Citation to the report describing this subsystem omitted due to double-blind constraints.}
More details can be found in~\cite{chato2013}. % ANON~\cite{castillo_2013_matching}.

% reduced the details on IntoNow -- gdfm
The matching is performed in two steps. In the first step, a per-genre classification model trained on thousands of examples labeled by editors is applied. In this model, the two classes are ``same story'' and ``different story'' and each example consists of a sentence, a news story, and a class label.
The features for the classifier are computed from each sentence-story pair by applying the named entity tagger described in the previous section on both elements of the pair, and then by looking at entity co-occurrences.
%The models are fine-tuned to have about 90\% precision, with recall in the 50-60\% range.
The models are fine-tuned to have high precision.

In the second step, recall is improved by
aggregating multiple sentence-level matchings that occur in a short time period to form a ``qualified matching''.% Unsure about whether to include this footnote --ChaTo: \footnote{The news matching system is a key asset of the company that provides our data, hence our description stays at the general level. The specific method for matching captions to news is not central to our contribution.

% !TEX root = paper.tex
\section{News Providers}\label{sec:providers}

In this section we examine news providers, and try to answer the following research questions:

\begin{question}[Styles and genres\label{q:stylistic-differences}]
Are there NLP-based attributes of newscasts that correlate with the genre of each provider?
\end{question}

\begin{question}[Coverage\label{q:coverage-differences}]
Do major networks have more coverage of news events compared to minor ones?
\end{question}

\begin{question}[Timeliness\label{q:temporal-differences}]
To what extent ``breaking'' a news story depends on covering a large number of stories?
\end{question}

\subsection{Styles and genres} \label{subsec:styles-genres}

We apply FABIA~\cite{fabia2010} to obtain a soft bi-clustering of news providers according to the words they use more frequently %; the number of clusters (ten) was determined by a tuning phase.
%ignoring for now named entities, which are analyzed in Section~\ref{sec:newsmakers}.
(we ignore for now named entities, which are considered in Section~\ref{sec:newsmakers}).
The output is shown in Table~\ref{tbl:lemma_provider_co_clustering}, where we have included descriptive providers and words for each cluster.

Most of the cohesive clusters are ``pure'', i.e., they have a single genre. Additionally, the three most popular networks are clustered together in the sixth cluster. % with the most audience~\cite{pew2012tv} \prov{Fox News}, \prov{MSNBC} and \prov{CNN} are clustered together.
Among the non-pure clusters, the third one is the most interesting.
\provider{E}{\aEntertainment}, an entertainment news service, and \provider{CNN Headln}{\aGeneral} cluster around descriptive words such as \emph{wearing}, \emph{love}, and \emph{looks}, terms strongly related to the world of \fEntertainment news.
While \provider{E}{\aEntertainment} is expected to use such words, its similarity to \provider{CNN Headln}{\aGeneral}
can be attributed to at least two factors: (i) a deliberate stylistic and content choice made to compete with other fast-paced headline-oriented providers, and (ii) intrinsic aspects of the headline format, which is less formal, less deep, and short in nature, which inevitably leads to a more superficial coverage.

Finally, the grouping of \fBusiness and \fSports providers at the bottom of Table~\ref{tbl:lemma_provider_co_clustering} is also of interest.
They use similar polysemic terms such as \emph{beats}, \emph{scoring}, \emph{wins}, \emph{loses}.
The use of a shared terminology stems from the common nature of competition associated with both \fSports and \fBusiness.

\begin{figure}[ht!]
\centering\includegraphics[width=.6\columnwidth, angle=-90]{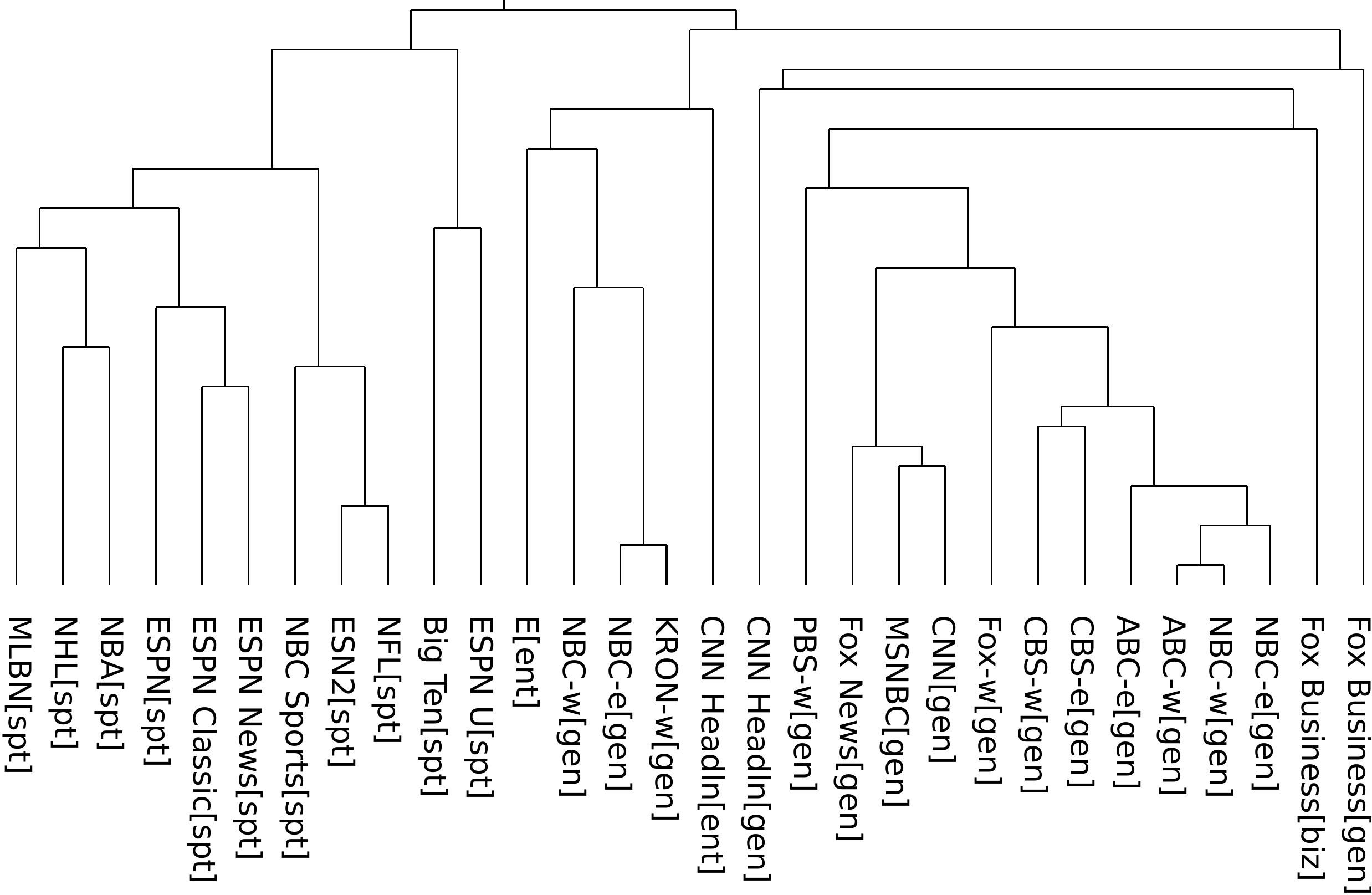}
\caption{Hierarchical clustering of providers based on the prevalence of different linguistic classes.}
\label{fig:channel_clusters_style}
\end{figure}

\spara{Part-of-speech classes and dependencies.}
We use part-of-speech and dependency tags to analyze the differences in style among providers.
%A previous work done on online news observed that, e.g., adjectives are more commonly used in {\em fashion} and {\em art} articles~\cite{flaounas2013digital_journalism}. Table~\ref{tbl:linguistic_classes} summarizes the main linguistic categories found in our dataset.
We represent each provider as a distribution over linguistic categories (e.g., number of verbs, number of adjectives), and apply hierarchical agglomerative clustering with euclidean distance to this representation.
Figure~\ref{fig:channel_clusters_style} shows the resulting clustering of the top-30 providers with most mentions.

\begin{table}[t]
\caption{Provider-word bi-clustering results. Clusters are sorted by cohesiveness (top to bottom), and the main descriptive providers and words for each cluster are shown. Clusters may overlap.}
\label{tbl:lemma_provider_co_clustering}
\centering\scriptsize
\begin{tabular}{lp{2.1in}}\toprule
{\bf Providers} & {\bf Words} \\\midrule
\provider{Fox News}{\aGeneral}        &  the, said, has, says, was, \\
\provider{CNN}{\aGeneral}             &  saying, former, do, get, made\\ \midrule
\provider{MLBN}{\aSports}            &  save, get, facing, was, goes \\
\provider{ESPN Classic}{\aSports}    &  hit, got, coming, run, getting, see \\
\provider{NBC-w}{\aSports}             &  out, here, catch, enjoys, leading \\ \midrule
\provider{E}{\aEntertainment}  &  wearing, have, was, it, has \\
\provider{CNN Headln}{\aGeneral}    &  had, think, love, looks, know \\ \midrule
\provider{ESN2}{\aSports}            &  later, taking, wins, rose, hitting \\
\provider{ESPN News}{\aSports}       &  get, that, got, looking, won, win\\ \midrule
\provider{NFL}{\aSports}          &  throw, suspended, free, be, get \\
\provider{ESPN News}{\aSports}      &  threw, played, said, one, traded \\ \midrule
\provider{Fox News}{\aGeneral}         &  reverend, vote, said, endorsed, voted \\
\provider{MSNBC}{\aGeneral}            &  saying, elected, support, attacking, running \\
\provider{CNN}{\aGeneral}              &  defeat, attack, wants, calling, conservative \\ \midrule
\provider{NBA}{\aSports}           &  finds, missing, knocking, scoring, shot \\
\provider{ESPN News}{\aSports}     &  taking, playing, later, finding, play \\
\provider{ESN2}{\aSports}           &  passing, driving, out, shoot, lays \\ \midrule
\provider{MSNBC}{\aGeneral}             &  save, stopped, played, gets, makes  \\ 
\provider{NHL}{\aSports}      &  helping, comes, playing, goes, ends \\
\provider{CNN Headln}{\aGeneral}     &  shut, one, traded, beats, scoring  \\ \midrule
\provider{Fox Business}{\aBusiness}   &  later, taking, finds, facing, wins \\ 
\provider{ESN2}{\aSports}            &  passing, beats, scoring, visiting, pitched \\   \midrule
\provider{Fox Business}{\aBusiness}   &  save, later, finds, taking, beats \\ 
\provider{NHL}{\aSports}      &  scoring, looking, hosting, stopped, scored \\ \bottomrule
\hline
\end{tabular}
\end{table}
 
\iffalse
\begin{table}[t!]
\caption{Main part-of-speech classes, including examples and total number of occurrences in our data.}
\label{tbl:linguistic_classes}
\scriptsize\centering\begin{tabular}{lllr}\toprule
Tag &	Example word & Class 	& Occurrences\\\midrule
DT &	the	& Determiner	& 88K\\
JJ &	large	& Adjective	& 188K\\
%JJR&	larger	& ... comparative & 7K\\
%JJS&	largest	& ... superlative & 4K\\
NN &	player & Noun		& 878K\\
NNS&	players & ... plural	& 169K\\
NNP&	Indiana & ... proper   & 156K\\
%NNPS&	Maldives& ... proper plural & 1K\\
PRP&	me	& Personal pronoun & 51K\\
RB &	fast   & Adverb		& 90K\\
%RBR &	faster & ... comparative		& 1K\\
%RBS &	fastest& ... superlative		& $<$1K\\
VB  &	show	& Verb		& 273K\\
VBD & 	showed  & ... past	& 297K\\
VBG &	showing & ... gerund	& 227K\\
VBN &	shown  & ... past participle & 141K\\
VBP &	have  	& ... 1st/2nd present & 139K\\
VBZ &	has 	& ... 3rd present     & 273K\\\bottomrule
\end{tabular}
\end{table}

\fi

The clustering presents three clear super-groups: \fSports news on the left, \fEntertainment news in the middle, and  \fGeneral and \fBusiness news on the right.
Thus, while \fBusiness providers share their vocabulary with \fSports providers, their style is closer to \fGeneral providers.

\prov{Fox News} and \prov{MSNBC} are often considered antagonistic organizations with polarizing conservative and liberal views.
However, from the perspective of style they are similar, and also similar to \prov{CNN}.
Therefore, the three most popular networks are similar both in their vocabulary and style.
One outlier is \prov{PBS}, essentially a public broadcaster whose style is quite different from the major networks.
Finally, both \prov{KRON} and \prov{NBC} (which are affiliates and share several programs) show stylistic similarities to \fEntertainment providers even when broadcasting \fGeneral news.

%NEW PAGE
%\clearpage

\begin{figure}[h!]
\centering\includegraphics[clip=true, trim=0 0 0 8, width=.9\columnwidth]{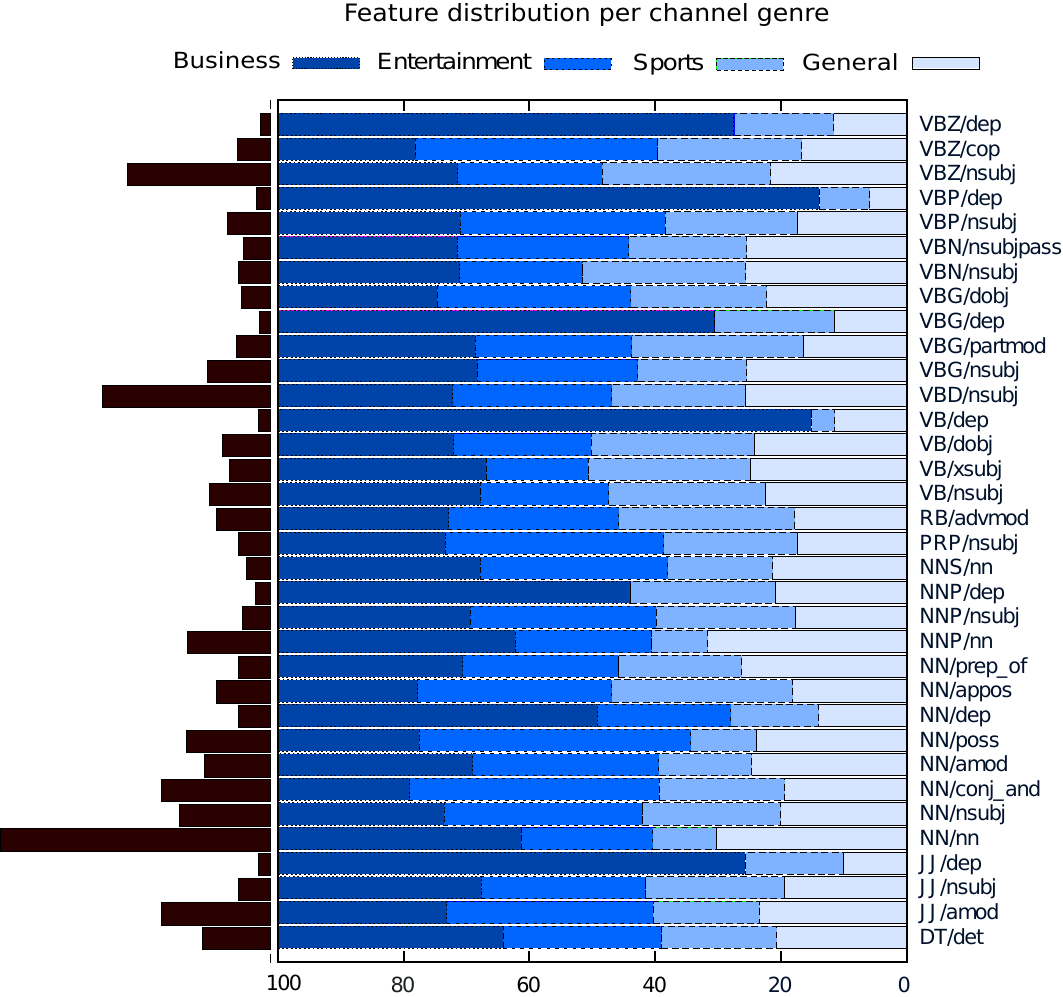}
\caption{Distribution of morphological and dependency types per provider genre.}
\label{fig:types_stack_style}
\end{figure}

Next we proceed to aggregate the linguistic categories at the level of genres.
Figure~\ref{fig:types_stack_style} presents the results, where we have also included the type of dependency found. 

In a large number of cases for \fBusiness providers, the dependency parser cannot extract the correct dependency (label ``\texttt{/dep}''), while for \fEntertainment providers the incidence is very small.
%This specific parser is trained on a combination of Wall Street Journal articles plus other technical and general documents.
A possible interpretation of this difference may be due to a different complexity of phrases.
As observed by~\citet{flaounas2013digital_journalism} for online news, politics, environment, science, and business use a more complex language; sports, arts, fashion and weather a simpler one.
The analysis of other variables in our data seems to support this hypothesis.
A typical sentence has $9.1$ words in \fSports news, $9.0$ words in \fEntertainment news, $11.5$ in \fGeneral news, and $13.2$ in \fBusiness news.
The Fog readability index~\citep{gunning1952technique} (which estimates the years of formal education needed to understand a text on a first reading) is  $6.7$ for \fSports, $7.2$ for \fEntertainment, $9.1$ for \fGeneral, and $9.4$ for \fBusiness.

%	Tokens	Chars	Percent of complex words	Average words per sentence	Fog	Flesch	Flesch-Kincaid
%sports	6.87618962348571	36.1171093734159	7.52	9.1197	6.6573	79.7697	4.3986
%entertainment	6.97172404262507	37.1883873824074	8.97	9.0087	7.1905	74.4095	5.1187
%news	7.54448148957424	41.2882062016484	11.34	11.4737	9.1251	68.0601	6.6167
%business	7.8561094810045	41.8731279375091	10.22	13.2035	9.3706	70.2810	6.7367

For the other linguistic categories, \fEntertainment has the largest relative prevalence of {\tt NN/poss} (singular common noun, possession modifier, such as {``Kristen Bell struggled to work with her \emph{fianc\'{e}}''}),
\fSports has the largest value for {\tt NN/appos} (singular common noun, appositional modifier, such as {``Kevin Love's 51 \emph{points}, a Minnesota Timberwolves team record''}),
and \fGeneral news has the largest value for {\tt NNP/nn} (singular proper noun, compound modifier, such as {``President \emph{Obama} is refocusing his campaign''}).

\begin{figure}[h!]
\centering\includegraphics[width=.63\columnwidth]{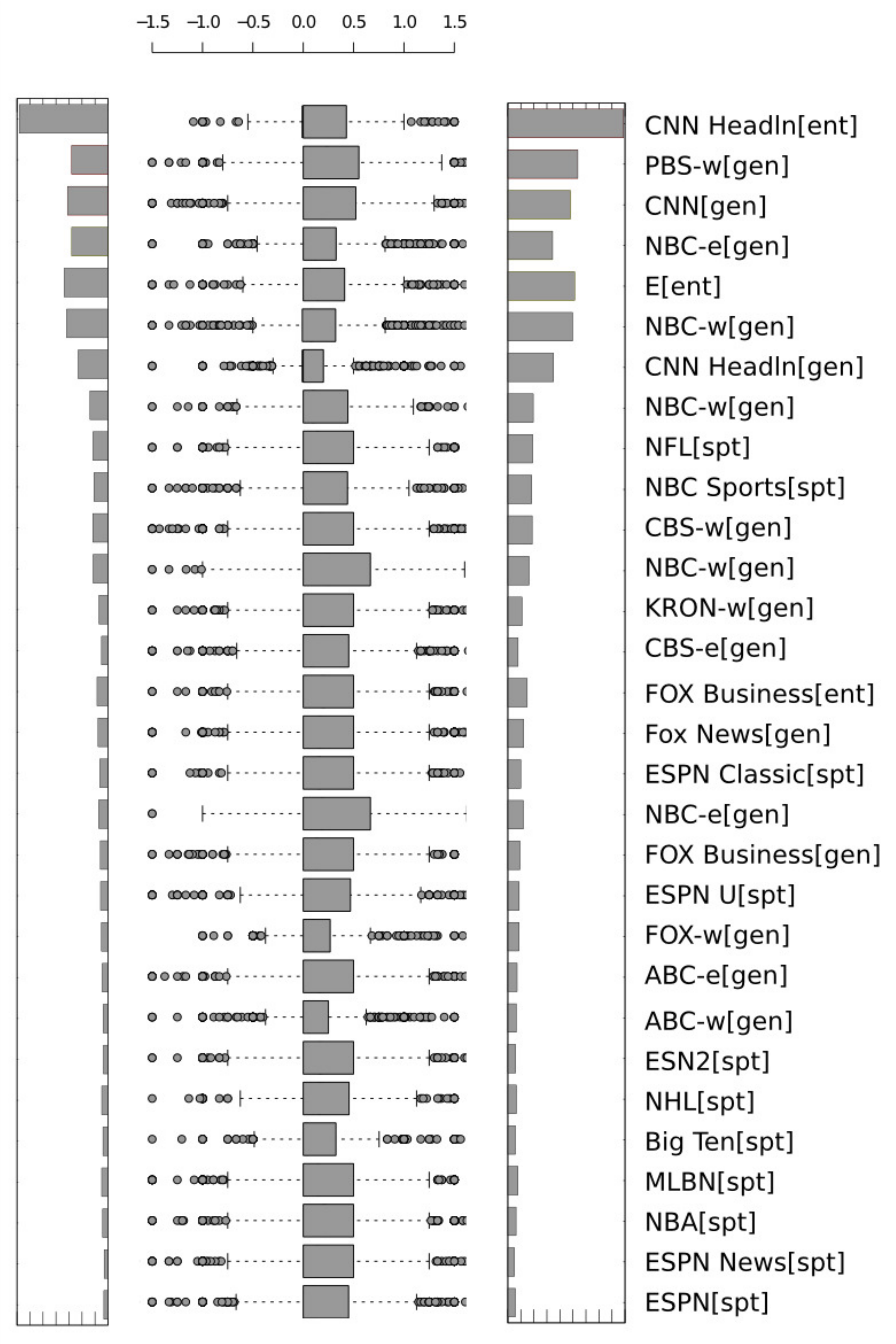}
\caption{Polarity distribution of sentences per provider, shown as box 
plots in the center of each line. Bars on the left and right  indicate the number of negative and positive words, respectively. Providers are sorted by average sentiment score, from  negative (top) to positive (bottom).}
\label{fig:polarity_per_channel}
\end{figure}

\spara{Sentiments.}
We analyze the distribution of sentiments expressed by each provider on each caption.
The result is shown in Figure~\ref{fig:polarity_per_channel}, in which we have included the number of negative words and positive words, as well as the distribution of sentiment scores.

The bulk of sentiments on news seem to range from neutral to positive.
All of the seven most positive providers are of the \fSports genre. \provider{CNN Headln}{ent} is an outlier in at least two senses: it is the most negative and polarized provider, and it has many sentiment-loaded words (e.g. it has more sentiment-loaded words than \provider{CNN Headln}{gen}, even when its constitutes the minority of programs in \prov{CNN Headln}). This can be attributed to the ``attention-grabbing'' needs of the headlines format in the case of \fEntertainment news.

\subsection{Coverage}\label{subsec:coverage}

In this section and in the next one, we make use of the news matchings described in Section~\ref{subsec:news_matching}.
A provider {\em covers} a story if it has at least one qualified matching for it. 

When measuring coverage, we have to consider that some news stories are more prominent than others. We denote by {\em prominence} the fraction of providers of a given genre that covers a story, so a story has prominence 1.0 if it is covered by all the providers of a genre -- which is quite rare.

\begin{figure}[h!]
\centering\subfigure[Probability that a provider of \fGeneral\ news covers a story as a function of its prominence.\label{fig:coverage-fraction-news}]{\includegraphics[width=0.9\columnwidth]{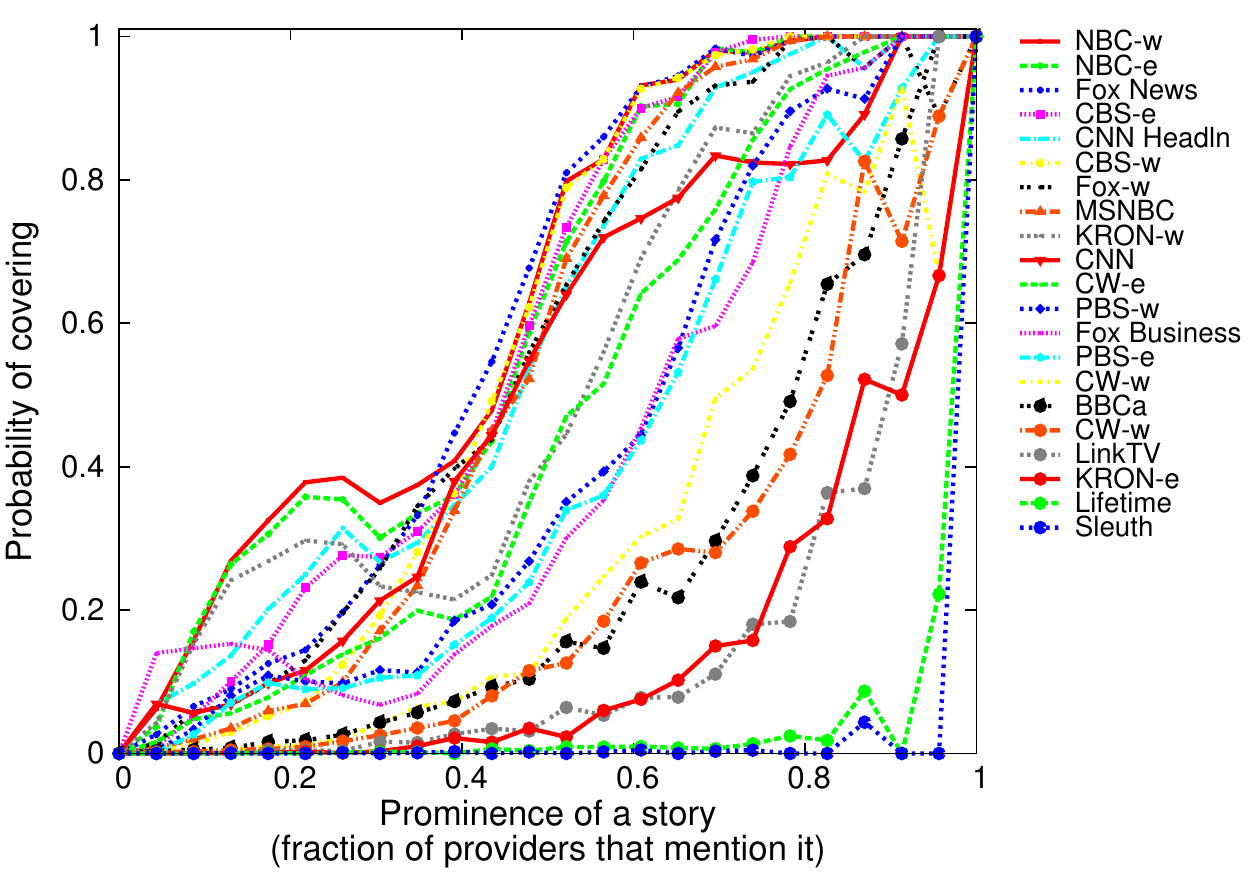}}
\subfigure[Distribution of prominence for \fGeneral\ news stories. The distribution is bimodal.\label{fig:coverage-number-news}]{\includegraphics[width=0.9\columnwidth]{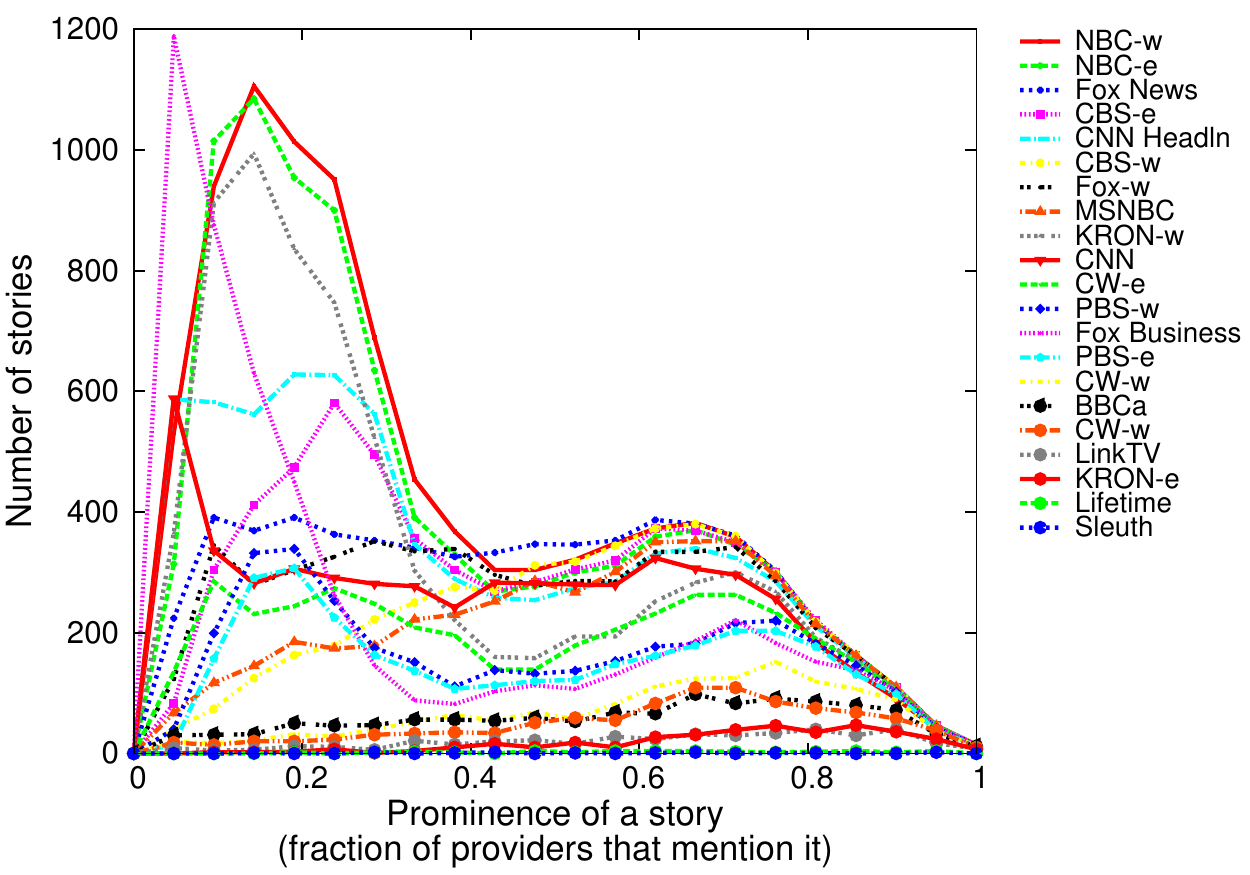}}
\caption{Relationship between coverage and prominence for different providers (best seen in color).}
\vspace*{-4mm}
\label{fig:coverage-consolidated}
\end{figure}

Figure~\ref{fig:coverage-fraction-news} shows the probability that a provider of \fGeneral\ news covers a story for different levels of prominence.
Some providers such as \prov{NBC}, \prov{Fox News}, \prov{CBS} and \prov{CNN Headln}, offer more extensive news coverage than others.
This wider selection of stories is likely due to having access to a larger pool of resources (e.g., employees, budget, affiliates) compared to the other \fGeneral news providers. 
This result is expected, however the data also suggests two other relevant findings.
\prov{NBC} and \prov{CNN Headln} seem to have a non-trivial coverage ($\approx 0.3-0.4$) of relatively niche stories (prominence $\approx 0.2$), content that is not covered by \prov{Fox News}. However, \prov{Fox News} has a wider coverage of stories having a prominence of $\approx 0.4$ and over, which means it reports on a higher number of stories than either \prov{NBC} or \prov{CNN}.

%This result profiles two distinct editorial approaches to covering the news, stemming from two distinct editorial agendas.
%\prov{NBC} -- which is often associated with a liberal agenda -- chooses to spend resources to cover niche topics and issues,
%while \prov{Fox News} -- favored by conservatives -- prefers to cover mostly wide-reaching, mainstream stories.
%adopts a wide reaching approach to news coverage; we see these are clearly contrarian entities in more ways than one.

Figure~\ref{fig:coverage-number-news} also shows coverage on \fGeneral news, this time from the point of view of the distribution across different levels of prominence.
The distribution is clearly bimodal, with the first mode around $3$, and the second one around $14$.
Most news are covered by just a handful of providers, while a few manage to catch the attention of many providers.

%\prov{Fox Business} has a large share of unique stories, probably due to the introduction of more niche stories from the \fBusiness domain, even in their \fGeneral\ news programs.
%In addition, a person watching \prov{NBC}, \prov{KRON}, and to some extent \prov{CNN Headln} has a relatively higher chance to be exposed to news that are not shown by many other providers.

\subsection{Timeliness and duration}\label{subsec:timeliness}

\begin{figure}[ht!]
   \centering
   \includegraphics[clip=true, trim=0 15 0 0, width=\columnwidth]{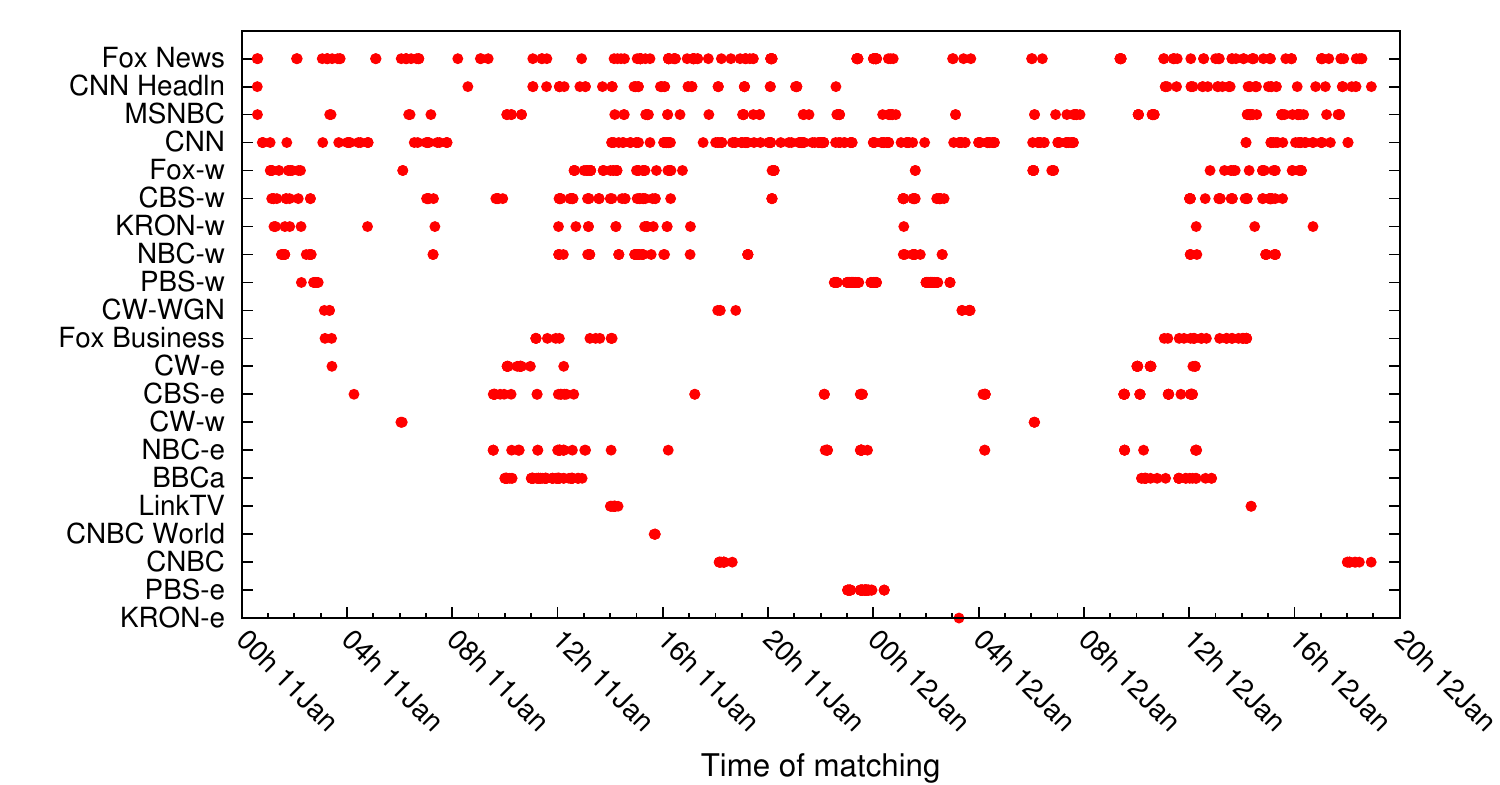}
   \caption{Example of matchings for a news story during two days on January 2012. The major channels are the fastest to break the news, and the minor ones follow them. The major providers have the highest density of matchings.}
   \label{fig:matching-example-news}
\end{figure}

In this section we examine how different providers cover a story over time. 
The life cycle of a prominent news story is exemplified by Figure~\ref{fig:matching-example-news}, which depicts the coverage of an abuse probe involving US marines during two days on January 2012.
Each dot represents a matching of this news story with a provider.

Providers are sorted by earliest matching: in this case, \provider{Fox News}{\aGeneral}, \provider{CNN Headln}{\aGeneral} and \provider{MSNBC}{\aGeneral} are the first to broadcast the story, within minutes of each other.
There are two time-dependent variables of interest that we can measure.
The most important one is how often a provider is among the first ones to broadcast a story (i.e., ``breaks'' a story).
The second is how long the providers keep broadcasting a given story.

\begin{figure}[b!]
   \centering
   \includegraphics[width=\columnwidth]{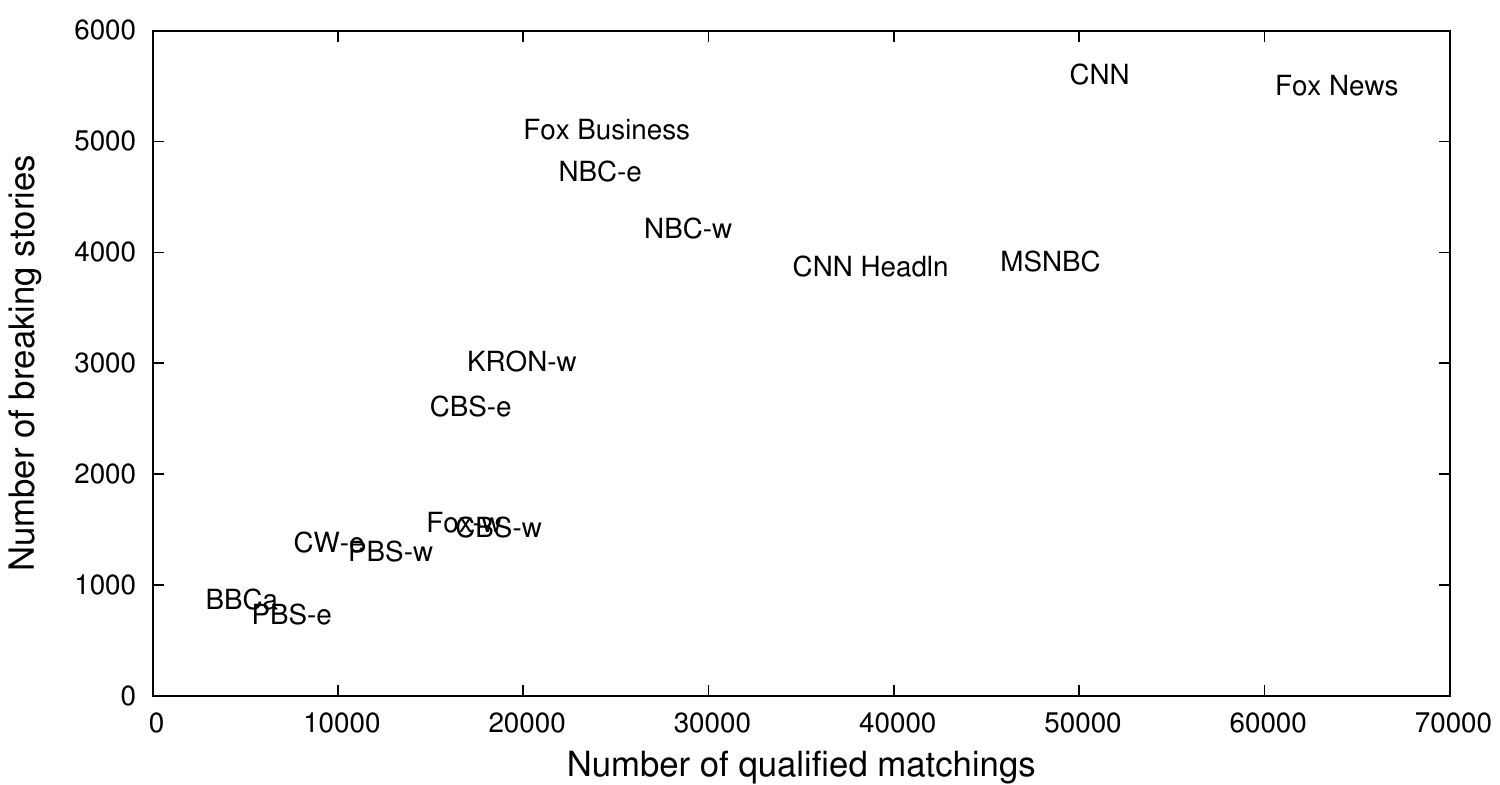}
   \caption{Number of breaking stories vs. number of qualified matchings, for providers of genre \fGeneral.}
   \label{fig:breaking-scatterplot-news}
\end{figure}

\spara{Timeliness.} Given that many news networks operate in cycles of 60 minutes, we consider that a provider ``breaks'' a story if it is among the ones that broadcast the story within the first 60 minutes since the first story matching.
%Figure~\ref{fig:breaking-scatterplot-news} compares the number of ``breaking'' stories with the number of qualified matchings for \fGeneral providers.
Figure~\ref{fig:breaking-scatterplot-news} plots providers \fGeneral news along two axes: how many qualified story matchings they provide and how many of those correspond to ``breaking'' a story.
Most of the providers lie along the diagonal (with a ratio of $\approx 1/10$), with the exception of \prov{Fox Business} and \prov{NBC}.
While these providers do not cover as many stories as \prov{CNN} and \prov{Fox News}, they are clearly better at breaking stories.

\spara{Duration.} We define the {\em duration} of a story as the interval between the first and the last matching.
This interval is bounded by the lifetime of the stories, which is three days.
Table~\ref{tab:duration} reports the average duration per provider.

% Requires the booktabs if the memoir class is not being used
\begin{table}[t]
   \scriptsize\centering
   \caption{Average duration of a story on the top-24 providers with the most qualified matchings, i.e. the time span between the first and the last matching (hours).}
   \begin{tabular}{@{} l r l r @{}} % Column formatting, @{} suppresses leading/trailing space
      \toprule
      Provider    & Duration (h) & Provider & Duration (h)\\
      \midrule
\provider{ESPN Classic}{\aSports}              &        26.6        &           \provider{CNBC}{\aBusiness}          &          13.0 \\
\provider{ESPN News}{\aSports}    &             24.5        &           \provider{Fox Business}{\aGeneral}            &         12.1 \\
\provider{NFL}{\aSports}                &               23.9        &                   \provider{MSNBC}{\aGeneral}          &          11.8 \\
\provider{ESN2}{\aSports}              &        23.0        &                   \provider{Fox News}{\aGeneral}    &             11.0 \\
\provider{NBA}{\aSports}                &               22.5        &                   \provider{CNN}{\aGeneral}              &        11.0  \\
\provider{NHL}{\aSports}                &               21.6        &                   \provider{E}{\aEntertainment}      &            10.9  \\
\provider{MLBN}{\aSports}              &        19.3        &                   \provider{CBS-w}{\aGeneral}          &          9.8    \\
\provider{NBC Sports }{\aSports}                &               18.2        &           \provider{KRON-w}{\aGeneral}        &           9.6    \\
\provider{ESPN U}{\aSports}          &          17.9        &                   \provider{NBC-e}{\aGeneral}          &          8.4     \\
\provider{ESPN}{\aSports}              &        17.4        &                   \provider{NBC-w}{\aGeneral}          &          8.4     \\
\provider{CNBC World}{\aBusiness}              &        15.7        &           \provider{CNN Headln}{\aGeneral}                &               7.6     \\
\provider{Bloomberg}{\aBusiness}                &               15.4        &           \provider{PBS-w}{\aGeneral}          &          5.7     \\
      \bottomrule
   \end{tabular}
   \label{tab:duration}
\end{table}

The longest duration is found in \fSports providers, followed by \fBusiness, then \fGeneral ones.
%We can observe that \fSports news tends to be a genre where stories have a long duration.
Indeed, a game that occurs over the week-end can be commented during several days.
For the major \fGeneral news providers, the typical duration of a story is from 8 to 12 hours.

% !TEX root = paper.tex
\section{Newsmakers}\label{sec:newsmakers}

This section presents an analysis of {\em newsmakers}, i.e., people who appear in news stories on TV. We consider the following research questions:

%\begin{question}[People and news]\label{q:people-news}
%Which persons are more prominently mentioned on television news, and how?
%\end{question}

\begin{question}[Newsmakers by profession]\label{q:newsmakers-professions}
Are there observable differences in the way different professions are covered and portrayed by news programs that can be detected by using our NLP-based pipeline?
\end{question}

\begin{question}[Newsmaker groups]\label{q:newsmakers-group}
To what extent can our NLP-based pipeline identify groups of similar newsmakers?
\end{question}

\subsection{Newsmakers by profession} \label{subsec:entity-role}

The named entity tagger we use~\cite{zhou2010resolving} resolves entities to Wikipedia pages, thus allowing us to obtain more information about people from those pages.
We scrape data from Wikipedia {\em infoboxes} to categorize newsmakers according to professional areas and activities, and obtain a coverage of $98.2\%$ of the mentions in our data.
Table~\ref{tbl:polarity_per_role} shows the 24 most mentioned professions. % with their average sentiment score, and up to three representative persons per profession.
The table spans a large range of prominence, from the most mentioned profession having more than $400$k mentions to the last one shown in the table having only $3$k.

\begin{table}[t]
\caption{Top-24 occupations with the most mentions, including average sentiment and example persons.}
\label{tbl:polarity_per_role}
\scriptsize\centering
\hskip-1em\begin{tabular}{@{}l@{}r@{\,\,\,\,}p{2.0in}}\toprule
Activity&Sent.&Most mentioned people \\\midrule
Sports/Basketball&\mbox{-0.05}& LeBron James 11\%, Kobe Bryant 5\%, Dwyane Wade 4\% \\
Sports/Football&0.06& Tim Tebow 12\%, Peyton Manning 9\%, Tom Brady 5\% \\
Politics/US GOP&\mbox{-0.09}& Mitt Romney 39\%, Newt Gingrich 17\%, Rick Santorum 13\% \\
Politics/US DEM&\mbox{-0.21}& Barack Obama 76\%, Hillary Clinton 9\%, Joe Biden 4\%\\
Art/Music&\mbox{-0.01}& Whitney Houston 14\%, Neil Young 12\%, Jennifer Lopez 4\%\\
Sports/Baseball&\mbox{-0.01}& Albert Pujols 4\%, Justin Verlander 3\%, Bryce Harper 3\%\\
Art/Actor&0.04& George Clooney 4\%, Kim Kardashian 4\%, Brad Pitt 3\%\\
Sports/Ice hockey&0.06 & Martin Brodeur 11\%, Jonathan Quick 5\%, Ilya Bryzgalov 4\% \\
Sports/Golf&\mbox{-0.02}& Tiger Woods 31\%, Phil Mickelson 8\%, Jack Nicklaus 8\% \\
Media/ Showbiz & \mbox{-0.10}& Rush Limbaugh 19\%, Nicole Polizzi 13\%, Al Sharpton 7\% \\
Other/Business& \mbox{-0.10} & Warren Buffett 21\%, Jim Irsay 12\%, Mark Zuckerberg 10\% \\ 
Sports/Racing& \mbox{-0.02} & Dale Earnhardt 13\%, Danica Patrick 9\%, Jimmie Johnson 8\% \\
Sports/Tennis& \mbox{-0.06} & Rafael Nadal 26\%, Novan Djokovic 25\%, Roger Federer 13\% \\
Media/Journalist& 0.11 & Matt Lauer 12\%, Wolf Blitzer 11\%, Ann Curry 6\% \\
Sports/Martial arts& \mbox{-0.29} & Nick Diaz 13\%, Nate Diaz 8\%, Wanderlei Silva 5\% \\
%Other/Religion& \mbox{-0.03} & Adam 55\%, Jesus 21\%, Muhammad 12\% \\
Art/Comedian & 0.07 & Stephen Colbert 10\%, Bill Maher 8\%, Jay Leno 8\% \\
Sports/Boxing & \mbox{-0.26} & Muhhamad Ali 25\%, Manny Pacquiao 23\%, Mike Tyson 15\% \\
Mil./SA & \mbox{-0.92} & Osama bin Laden 100\%\\
Official/US & \mbox{-0.50} & Leon Panetta 43\%, Eric Holder 26\%, Jay Carney 13\% \\
Other/Other & \mbox{-0.22}  & Hilary Rosen 8\%, Jeremiah Wright 8\%, Andrea Yates 7\% \\
Sports/Soccer & 0.01 & Lionel Messi 40\%, David Beckham 27\%, Wayne Rooney 13\% \\
Art/Writer & \mbox{-0.07} & Ernest Hemingway 17\%, Andrew Breitbart 10\%, William Shakespeare 
7\% \\
Sports/Wrestling & 0.62 & Dwayne Johnson 81\%, Brock Lesnar 9\%, Hulk Hogan 6\% \\
Politics/US D Spouse & 0.02 & Michelle Obama 100\% \\
\bottomrule
\end{tabular}
\end{table}

\iffalse
The five most prominent professions in the news are basketball player, football player, Republican Party politician, Democratic Party politician, and musician. 
The rest of the list is dominated by other sportspeople, artists, and entertainers.
The relative prominence of different sports probably varies during the year: our observation period covers the key games of the basketball and football season, but only the initial part of the baseball season. Further down the list, businesspeople appear at the 11th place, journalists at the 14th place, and government officials at position 19.
\fi
\spara{Concentration of mentions per profession.}
The distribution of mentions per person in each profession varies substantially.
Politicians and government officials are represented by a few key individuals who attract the majority of mentions, which is consistent with the findings of \citet{schoenbach_2001_politicians}.
Our dataset spans across the US presidential campaign period, which may cause mentions to be even more concentrated around the top candidates.
A high level of concentration of mentions is observed also in businesspersons, dominated by Warren Buffett, and in individual sports such as golf, tennis, and wrestling.

\iffalse

With the exception of soccer (in which two-thirds of the mentions go to Lionel Messi and David Beckham), in team sports we observe that mentions are distributed across a wider range of athletes: this is the case for basketball, football, baseball, and hockey.
Another profession in which the distribution of mentions is comparatively less skewed is acting, with George Clooney (the most mentioned actor during this period) attracting a mere 4\% of mentions. Indeed, in recent years an increasing fragmentation of viewership of films has been reported\footnote{\scriptsize\texttt{http://www.nytimes.com/2012/10/29/movies/hollywood-seeks-to-slow-cultural-shift-to-tv.html}} with large Hollywood films facing increasing competition from TV films, dramas, and series. 

\fi

\spara{Sentiments.}
We first focus on individuals and select those that have at least 10k mentions.
The persons most associated with \emph{negative} words are: Osama bin Laden ($-0.92$), Whitney Houston (who passed away during the observation period, $-0.25$), and George W. Bush ($-0.21$).

The most associated with \emph{positive} words are three football stars: Andrew Luck\footnote{\scriptsize{This is likely to be to some extent, but not entirely, an artifact of ``luck'' being in the dictionary of the sentiment analysis software used.}} ($1.7$), Eli Manning ($0.24$) and Peyton Manning ($0.11$). 

In terms of professions, Democratic Party politicians get a more negative treatment than Republican Party politicians.
We observe that while the former are incumbent in the US government, the latter were undergoing their presidential primary during the first four months of our study. At each primary or caucus (there were tens of them) a number of winners or groups of winners were declared.

Overall, the most positive average score (0.62) is attained by professional wrestlers.
Note that Dwayne ``The Rock'' Johnson and '80s popular culture icon Hulk Hogan also have an important career as entertainers, with ``The Rock'' staging a much-publicized come back in early 2012.
The second most positive sentiment (0.11) is attained by journalists, thus indicating that they often refrain from criticizing or speaking in negative terms on air about their colleagues.

\subsection{Automatic clustering of newsmakers}\label{subsec:entity-clustering}

\iffalse
In Section~\ref{subsec:styles-genres} we showed that the distribution of linguistic classes can be used to cluster news providers.
We attempt the same with newsmakers, however, the resulting clustering (omitted due to space constraints) contains a mixture of homogeneous and heterogeneous groups. 
To a large extent, this can be explained by variations across providers, which we treat more in-depth in Section~\ref{sec:biases}.
\fi
%Instead,
Table~\ref{tbl:entities_clustered_per_provider} shows a clustering based on linguistic attributes for each of the top providers per genre.
%the same type of clustering broken down by provider, with the top providers per genre.
%Even though many of the entities in Table~\ref{tbl:clustering_entities} are clustered in the same way in Table~\ref{tbl:entities_clustered_per_provider}, there are nevertheless some differences.
Interestingly, \fEntertainment and \fSports programs tend to conflate all politicians in one cluster, whereas \fBusiness and \fGeneral providers tend to separate them.
For instance, \provider{CNN Headln}{\aGeneral} generates a clear cluster with all the primary candidates, and \provider{Fox News}{\aGeneral} separates primary candidates from final presidential candidates. This would suggest that \prov{Fox News} has a more nuanced coverage of Republican Party politics than the other networks.
Along the same lines, \provider{Fox Business}{\aBusiness} refers to a mixture of entertainment people (George Clooney, Kim Kardashian) and sports people (LeBron James, Kobe Bryant) using a similar style, while \provider{E}{\aEntertainment} exhibit stylistic differences in the way it speaks about male celebrities (George Clooney, Justin Bieber) and female ones (Lindsay Lohan, Britney Spears): \prov{E} displays a greater nuance in covering celebrity news.

In general, these differences suggest that providers are more discerning when covering people in their area of expertise, than when speaking about people outside it.

\begin{table*}[htdp]
\scriptsize\centering
\caption{Stylistic clustering per provider. Clusters are sorted by internal similarity, in decreasing order from top to bottom.}
\label{tbl:entities_clustered_per_provider}
\begin{tabular}{l|l|l|l|l|l}\toprule
\provider{ESPN}{\aSports} & \provider{Fox News}{\aGeneral} & \provider{CNN Headln}{\aGeneral} & \provider{Fox Bus.}{\aBusiness} & \provider{E}{ent} & \provider{KRON}{ent} \\\hline
LeBron James	&	Neil Young	&	Dwyane Wade	&	Mitt Romney	&	Jimmy Carter	&	Mitt Romney	\\
\cline{2-2} \cline{5-5}
Tim Tebow	&	Newt Gingrich	&	Kobe Bryant	&	Newt Gingrich	&	Lindsay Lohan	&	Newt Gingrich	\\
Peyton Manning	&	Rick Santorum	&	Joe Biden	&	Rick Santorum	&	Britney Spears	&	Tom Brady	\\
\cline{3-3} \cline{5-5}
Dwyane Wade	&	Ron Paul	&	Blake Griffin	&	Ron Paul	&	Bill Clinton	&	Bobby Brown	\\
\cline{2-2}
Kobe Bryant	&	Barack Obama	&	Carmelo Anthony	&	Bill Clinton	&	Brad Pitt	&	Matt Lauer	\\
\cline{3-3} \cline{5-6}
Jeremy Lin	&	Mitt Romney	&	Tom Brady	&	Neil Young	&	Mike Tyson	&	Whitney Houston	\\
\cline{2-2} \cline{5-5}
Kevin Durant	&	Whitney Houston	&	Drew Brees	&	Ronald Reagan	&	Eli Manning	&	Oprah Winfrey	\\
Tiger Woods	&	Bill Clinton	&	Derrick Rose	&	Joe Biden	&	Hillary Clinton	&	George Clooney	\\
\cline{1-1} \cline{4-4}
Neil Young	&	Ronald Reagan	&	Eli Manning	&	Barack Obama	&	Matt Lauer	&	Kim Kardashian	\\
\cline{1-1} \cline{4-4}
Barack Obama	&	Joe Biden	&	Ronald Reagan	&	Tim Tebow	&	Michael Jackson	&	Justin Bieber	\\
\cline{5-6}
Bill Clinton	&	John McCain	&	Chris Bosh	&	Eli Manning	&	Whitney Houston	&	Jimmy Carter	\\
\cline{3-3}
Eli Manning	&	John Edwards	&	LeBron James	&	Jimmy Carter	&	Tony Romo	&	Brad Pitt	\\
\cline{2-2}
Bill Belichick	&	LeBron James	&	Tim Tebow	&	John McCain	&	Jennifer Hudson	&	Lindsay Lohan	\\
\cline{1-1}
Jimmy Carter	&	Dwyane Wade	&	Peyton Manning	&	John Edwards	&	John Travolta	&	Britney Spears	\\
\cline{1-1} \cline{5-5}
Mitt Romney	&	Tom Brady	&	Jeremy Lin	&	Donald Trump	&	Sarah Palin	&	Will Smith	\\
\cline{5-5}
Newt Gingrich	&	Drew Brees	&	Whitney Houston	&	John Boehner	&	Barack Obama	&	Sarah Palin	\\
\cline{4-4}
Joe Biden	&	Martin Brodeur	&	Tiger Woods	&	Hillary Clinton	&	Mitt Romney	&	Rick Santorum	\\
\cline{4-4}
Ronald Reagan	&	Andrew Luck	&	Jimmy Carter	&	Tiger Woods	&	Dwyane Wade	&	Jeremy Lin	\\
\cline{1-1}
Whitney Houston	&	Eli Manning	&	John McCain	&	Sarah Palin	&	Kobe Bryant	&	John McCain	\\
\cline{1-1} \cline{3-4}
Joe Biden	&	Tiger Woods	&	Barack Obama	&	LeBron James	&	Jeremy Lin	&	John Edwards	\\
\cline{1-1}
Tom Brady	&	John Elway	&	Mitt Romney	&	Peyton Manning	&	Oprah Winfrey	&	Hillary Clinton	\\
Drew Brees	&	Mark Sanchez	&	Newt Gingrich	&	Dwyane Wade	&	Donald Trump	&	Donald Trump	\\
\cline{2-2}
Derrick Rose	&	Alex Smith	&	Rick Santorum	&	Tom Brady	&	Magic Johnson	&	Michelle Obama	\\
\cline{2-2}
Martin Brodeur	&	Sean Payton	&	Ron Paul	&	Kobe Bryant	&	Muhammad Ali	&	Mark Zuckerberg	\\
\cline{2-3}
Dwight Howard	&	Tim Tebow	&	Martin Brodeur	&	Jeremy Lin	&	Will Smith	&	Ann Romney	\\
\cline{5-5}
Chris Paul	&	Peyton Manning	&	Bill Clinton	&	John Elway	&	George Clooney	&	Charles Barkley	\\
\cline{3-3} \cline{6-6}
Blake Griffin	&	Kobe Bryant	&	Dwight Howard	&	Mark Sanchez	&	Justin Bieber	&	Tim Tebow	\\
\cline{5-5}
Chris Bosh	&	Jeremy Lin	&	Kevin Durant	&	Oprah Winfrey	&	Tim Tebow	&	Jennifer Hudson	\\
\cline{1-1} \cline{6-6}
Andrew Luck	&	Jimmy Carter	&	Andrew Luck	&	George Clooney	&	Tom Brady	&	Barack Obama	\\
\cline{6-6}
Carmelo Anthony	&	Michael Jordan	&	Bill Belichick	&	Kim Kardashian	&	Bobby Brown	&	Betty White	\\\bottomrule
\end{tabular}
%\note[ChaTo]{@Marcelo [FORMAT]: please reduce to only the first 7 columns, as this is the smallest font size at which this table makes sense, and it can only fit 7 columns of data. Marcelo wrote: I picked the top-providers by genre, 2 for news, 2 for ent, 1 for sports and 1 for bz, due to lack of space I can't pick 7 or 8 cols}
\end{table*} 

\begin{figure}[htb]
\centering\includegraphics[clip=true, trim=0 0 0 0, width=.92\columnwidth]{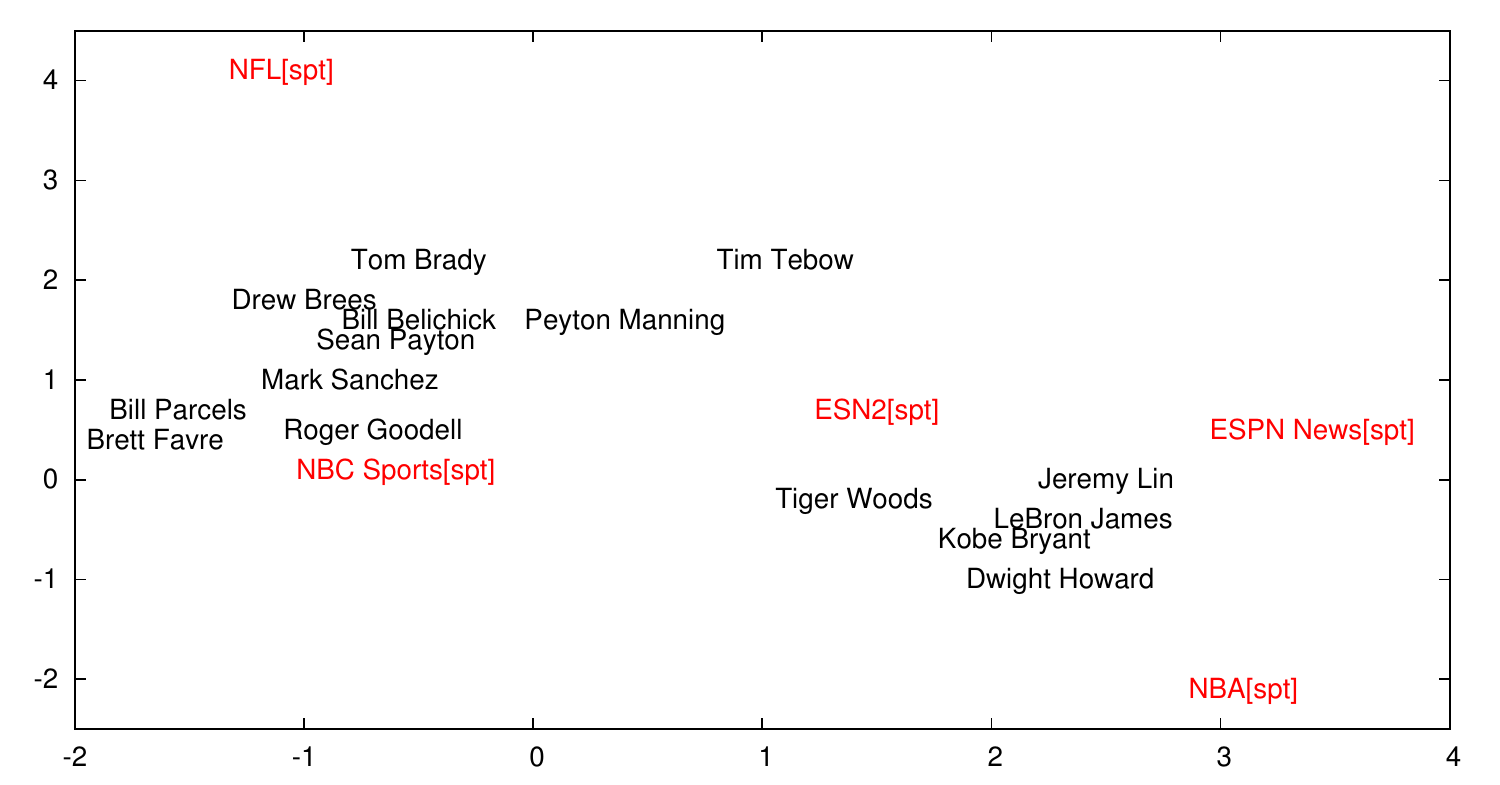}
\centering\includegraphics[clip=true, trim=0 0 0 0, width=.92\columnwidth]{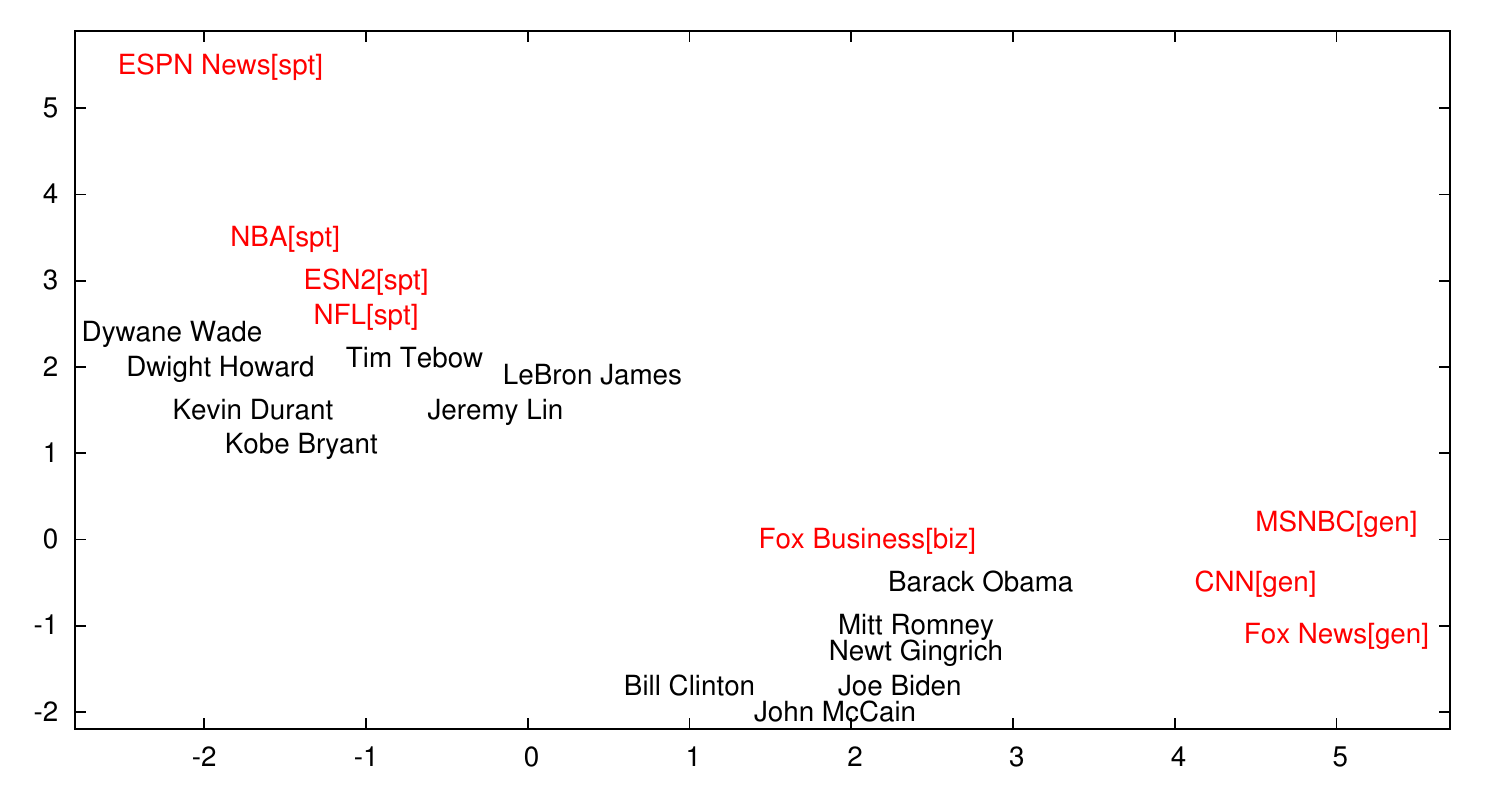}
\caption{Provider-newsmaker projections. The first component (top) separates football from basketball players, the second component (bottom), sportspeople from politicians.}
\label{fig:ac_1_2}
\end{figure}

\spara{Tensor decomposition.}
We further explore the stylistic relationship between newsmakers and news providers via a multi-way analysis of the principal components extracted from newsmaker-tag-provider triads. Figure~\ref{fig:ac_1_2} shows the result of projecting a three-way interaction model on two dimensions while fixing the linguistic tags.

This model is obtained by a three-way decomposition~\cite{giordani3}, which estimates a core tensor of the data. 
This technique is a natural extension of principal component analysis (PCA) for two-dimensional matrices. %<- Removable
We find that the dimensionality of the data is 3 (for the newsmakers dimension), 2 (for the linguistic dimension), and 3 (for the providers dimensions), and the tensor decomposition achieves a $78\%$ accuracy in 
reconstructing the original data. 

%We have a biplot projection of this three-way interaction model, by fixing POS-tags and projecting the relationships between 
%Channels and Entities through POS-tags. These plots are shown in Figures~\ref{fig:ac_1} and \ref{fig:ac_2} for the first and second component, respectively.

The first component neatly separates football from basketball players, which are the two most prominent professions in our dataset.
The sport-specific providers \provider{NBA}{\aSports} and \provider{NFL}{\aSports} appear near the axes, as naturally they cover more in depth their main sport.
Generalist \fSports news providers such as \provider{ESPN News}{\aSports} and \provider{ESN2}{\aSports} appear towards the top-right corner,
while \provider{ESPN News}{\aSports} seems to have a slight bias towards basketball.

The second component clearly separates sportspeople from politicians (the second most mentioned area in our dataset), together with the providers that mention each the most.
%It is interesting to note how the most representative entities for each category are able to be picked out from the rest of the mass.

%\note[gdfm]{ugly}
%Also worth noting that channels present a stronger ``bias'' on average compared to entities, which might suggest a constant style when referring to entities. \note[gdfm]{is it true?}\note[ChaTo]{Highly speculative.}

% !TEX root = paper.tex
\section{Framing and bias}\label{sec:biases}

\iffalse
In an often-cited definition, \citet{entman_1993_framing} defines \emph{framing} as selecting ``some aspects of a perceived reality and make them more salient in a communicating text, in such a way as to promote a particular problem definition, causal interpretation, moral evaluation, and/or treatment recommendation for the item described.''

Editors must frame for the benefit of their audience, in order to maximize the information they obtain during their attention windows.
\emph{Bias}, on the other hand, corresponds to presenting a partial perspective on facts~\cite{reese1996mediating},
or being perceived as being partial~\cite{niven_2003_bias}. Viewers are likely to perceive bias that is opposite to their partisan or ideological views~\cite{morris_2007_slanted}: liberals tend to find a conservative bias in media, while conservatives tend to find the opposite~\cite{coe_2007_partisan}.
\fi

In this section we analyze how different news providers frame different people. Specifically, we focus on the following research questions:

\begin{question}[Positive vs negative coverage]\label{q:biases-sentiment}
Are there significant differences in the sentiment polarity used by different providers when mentioning a given person?
\end{question}

\begin{question}[Outliers]\label{q:biases-outliers}
Are there people who are treated differently in a given provider compared to the rest?
\end{question}

\iffalse
\begin{question}[Professionals or entertainers]\label{q:biases-entertainers}
Can we identify athletes or politicians that appear in the news because of their celebrity status, independently of their professional merit?
\end{question}
\fi

\subsection{Positive vs negative coverage per provider}

Table~\ref{tbl:sent_score_entity_channel} shows the average sentiment of the four most mentioned people (two politicians and two sportsmen) across the providers with the largest number of mentions of people.
Previous works based on manual coding have observed clear variations in the coverage of different candidates by the major networks~\cite{morris_2005_news}.
Our automatic analysis confirms this finding: while Obama and Romney are treated equally by \prov{CNN} and \prov{MSNBC}, \prov{CNN Headln} and \prov{Fox News} give Romney a more positive treatment.
An even larger difference favoring Romney is exhibited by \prov{Fox Business}.

With respect to \fSports news, we notice an interesting phenomenon.
\prov{NBA}, specialized in basketball, speaks more positively about the football player (Tim Tebow);
conversely, \prov{NFL}, specialized in football, speaks more positively about the basketball player (LeBron James).
On average, Tim Tebow receives a more positive coverage than LeBron James, who among other things is still criticized for a team transfer in 2010, and according to USA Today became in 2012 one of the most disliked athletes in the US.\footnote{\url{http://usat.ly/xgiJ25}}
%http://content.usatoday.com/communities/gameon/post/2012/-02/vick-tiger-lead-list-of-americas-most-disliked-athletes-michael-vick-tiger-woods-forbes/1 %I AM USING THE SHORTENER OF USA TODAY, WHICH SHOULD BE AS STABLE AS USA TODAY URLS

\begin{table}[tbp]
\caption{Average sentiment scores of the 4 most mentioned persons in the top-15 providers with most mentions, grouped by type of provider. Empty cells mean no mentions.}
\label{tbl:sent_score_entity_channel}
\centering\scriptsize
\begin{tabular}{lcccc} \toprule
\multirow{2}{*}{Provider}	&	Barack 	&	Mitt 		&	LeBron 	&	Tim 	 \\
					&	Obama	&	Romney	&	James	&	Tebow	 \\ \midrule
\provider{Fox News}{\aGeneral}	&	0.24	&	0.33	&	0.28	&	0.37	\\
\provider{MSNBC}{\aGeneral}		&	0.28	&	0.28	&	-0.02	&	0.15	\\
\provider{CNN}{\aGeneral}		&	0.30	&	0.30	&	0.01	&	0.44	\\
\provider{CNN Headln}{\aGeneral}	&	0.22	&	0.35	&	0.14	&	0.25	\\ \midrule
\provider{ESPN News}{\aSports}	&	0.10	&	0.20	&	0.06	&	0.15	\\
\provider{NFL}{\aSports}		&	0.66	&	-	&	1.22	&	0.20	\\
\provider{ESN2}{\aSports}		&	0.46	&	0.14	&	0.13	&	0.20	\\
\provider{NBA}{\aSports}	&	0.21	&	-0.08	&	0.06	&	0.14	\\
\provider{NHL}{\aSports}	&	0.08	&	-	&	0.08	&	0.37	\\
\provider{NBC Sports}{\aSports}	&	0.48	&	0.30	&	0.24	&	0.32	\\
\provider{ESPN Classic}{\aSports}	&	0.25	&	-	&	0.12	&	
0.15	\\
\provider{MLBN}{\aSports}	&	-0.04	&	0.25	&	0.10	&	0.33	\\
\provider{ESPN}{\aSports}	&	0.18	&	-	&	0.07	&	0.04	\\ \midrule
\provider{Fox Business}{\aBusiness}	&	0.18	&	0.31	&	0.32	&	0.22	\\ \midrule
\provider{E}{\aEntertainment}		&	0.34	&	0.37	&	-	&	0.18	\\ \bottomrule
\end{tabular}
\end{table}

\subsection{Outlier analysis}

%We perform an analysis of outliers based on POS-based attributes (table omitted due to space limitations), and we find that the top outliers in each provider tend to include people that are famous in areas not usually covered by the provider, similarly to Table~\ref{tbl:entities_clustered_per_provider}.
%Instead, we present in Table~\ref{tbl:outlier_entities_per_provider_by_lemmas} the result of extracting for each provider the top people for which the distribution of words in the sentences mentioning them differs the most from the background distribution for each provider.

%Instead,
We show outliers by vocabulary in Table~\ref{tbl:outlier_entities_per_provider_by_lemmas}.
For each provider, we present the people whose distribution of words in sentences mentioning them differs the most from the background distribution.

With the exception of \provider{Fox News}{\aGeneral}, other \fGeneral news providers use a more specific vocabulary when speaking about current US president Barack Obama. % using a vocabulary that is different from the other people they cover. 
Interestingly, for \provider{ESPN News}{\aSports} and \provider{KRON}{\aEntertainment} the most significant outliers (Junior Seau and Whitney Houston, respectively) passed away during the observation period.

\begin{table}[h!]
\caption{Top five outliers per channel according to the distribution of words in their mentions.}
\label{tbl:outlier_entities_per_provider_by_lemmas}
\centering\scriptsize\begin{tabular}{ccc}\toprule
\provider{CNN Headln}{\aGeneral} &       \provider{E}{\aEntertainment}  &       \provider{ESPN News}{\aSports} \\
\midrule
Barack Obama    &       Britney Spears    &       Junior Seau     \\
Mitt Romney         &       Whitney Houston    &       Hines Ward  \\
Newt Gingrich         &       George Clooney      &       Ryan Braun       \\
Whitney Houston  &       Tim Tebow     &       Jerry Jones    \\
Tim Tebow     &       Justin Bieber       &       Joe Philbin        \\
\midrule
\provider{ESPNU}{\aSports}  &    \provider{Fox Business}{\aGeneral}      &       \provider{Fox News}{\aGeneral} \\
\midrule
John Calipari        &       Barack Obama    &       Jimmy Carter     \\
Brittney Griner      &       Mitt Romney      &       Mitt Romney      \\
Andrew Luck    &       Rick Santorum     &       Whitney Houston      \\
Rick Pitino     &       Ron Paul   &       Barack Obama     \\
Jared Sullinger   &       Newt Gingrich     &       Dianne Feinstein    \\
\midrule
\provider{KRON}{\aEntertainment} & \provider{KTVU}{\aGeneral} & \provider{NFL}{\aSports} \\
\midrule
Whitney Houston & Barack Obama       & John Elway \\
Oprah Winfrey & Mitt Romney     & Brett Favre \\
Britney Spears & Newt Gingrich        & Joe Montana \\
John Travolta & Rick Santorum      & Eli Manning \\
Jessica Simpson & Ron Paul & Sam Bradford \\
\bottomrule
\end{tabular}
\end{table}

\section{Conclusions} \label{sec:conclusions}

New domains provide both new challenges for NLP methods and new insights to be drawn from the data.
%The widespread availability of new text collections has invariably affected the research in Information Retrieval.
%Data from the World Wide Web propelled the field forward in the 1990s, and data from social media gave it a new boost in the 2000s.
Closed captions data for television programs, now available from the Internet Archive, brings the television domain within the realm of data mining research.
%This data has the potential to bring the order from the IR community to the subjectiveness of media experts.
%Objective evidence can now be collected to confirm or refute the subjective judgments of media experts.
%Automatic transcripts of speech are more precise today than ever.
%Researchers and practitioners are already designing and building applications in this new domain.
As already happened with the Web, we expect that mining this new source of data will provide a variety of interesting results.

% Deep characterization of new collection
%The effectiveness of operations such as source selection, document indexing and relevance ranking depends to a large extent on understanding deeply the collections over which such operations are applied.
%Users of television retrieval systems may be quite familiar with the biases in the collection and may expect, for instance, that the results include political and ideological diversity.

We outlined the main results of an ambitious study on this large collection of closed captions, focusing on the domain of news. 
%We applied the features extracted automatically by our pipeline to the task of characterizing news providers, newsmakers and their interaction.
%We analyzed the dataset from different points of view, and 
We demonstrated the richness of this dataset by analyzing several aspects such as the relationship between genres and styles, coverage and timeliness, and sentiments and biases when mentioning people in the news.
%From our analysis we can conclude that stylistic classes are strongly associated with types of news.
% Universal pipeline
The NLP-based pipeline proposed in this paper breaks the stream of text into sentences, extracts entities, annotates the text with part-of-speech tags and sentiment scores, and finally finds matching online news articles.
These steps provide the building blocks for any future task that leverages closed caption data, e.g., stream segmentation in topically homogeneous and semantically meaningful parts, classification and clustering of television news or characterization of people and events.

\section{Acknowledgments}
The authors wish to thank Ajay Shekhawat from Yahoo!
Marcelo Mendoza was supported by project FONDECYT grant 11121435. Carlos Castillo did most of this work while he was in Yahoo! Research.

\bibliographystyle{nourlabbrvnat}
\bibliography{mnlp}

\end{document}